\documentclass{article}

\usepackage[preprint]{neurips_2026}

\usepackage[dvipsnames]{xcolor}
\usepackage[utf8]{inputenc} % allow utf-8 input
\usepackage[T1]{fontenc}    % use 8-bit T1 fonts
\usepackage[pagebackref=true,breaklinks,colorlinks,citecolor=blue,linkcolor=blue,urlcolor=blue,hypertexnames=false]{hyperref}
\usepackage{url}            % simple URL typesetting
\usepackage{booktabs}       % professional-quality tables
\usepackage{amsfonts}       % blackboard math symbols
\usepackage{nicefrac}       % compact symbols for 1/2, etc.
\usepackage{microtype}      % microtypography
\usepackage{xcolor}         % colors
\usepackage{graphicx} 
\usepackage{siunitx}
\usepackage{amsmath}
\usepackage{amssymb, amsfonts}
\usepackage[accsupp]{axessibility}
\usepackage{inconsolata}
\usepackage{multirow}
\usepackage{tabularx}
\usepackage{colortbl}
\usepackage{pifont}
\usepackage{multicol}
\usepackage{booktabs}
\usepackage{makecell}

\usepackage{placeins}
\usepackage{marvosym}
\usepackage{utfsym}
\usepackage{xspace}
\usepackage{pifont}
\usepackage{arydshln}
\usepackage[T1]{fontenc}
\usepackage{sidecap}
\usepackage{enumitem}
\usepackage{lipsum}

\usepackage{array} 
\usepackage{threeparttable}
\usepackage{algorithm}
\usepackage{algorithmic}

\newcommand{\ourmethod}{ChronoWorld}

\title{\ourmethod{}: Camera-Controlled Consistent \\ 4D World Generation via Spatiotemporal Cues \\ and Geometric Reflections}

\author{
~ Xiaoyu Zhou\textsuperscript{1}
~ Dingwei Xian\textsuperscript{1}
~ Zhenyu Wang\textsuperscript{1} \\
~ \textbf{Yajiao Xiong\textsuperscript{1}}
~ \textbf{Yongtao Wang\textsuperscript{1,2}}\thanks{Corresponding author.}
~ \textbf{Ming-Hsuan Yang\textsuperscript{3}} \\
{\textsuperscript{1}Wangxuan Institute of Computer Technology, Peking University} \\
~ {\textsuperscript{2}VGI Labs Co., Ltd.}
~ {\textsuperscript{3}University of California, Merced} \\
}

\begin{document}

\maketitle

{%
\begin{figure}[h]
\hsize=\textwidth
\centering
\includegraphics[width=1.0\linewidth]{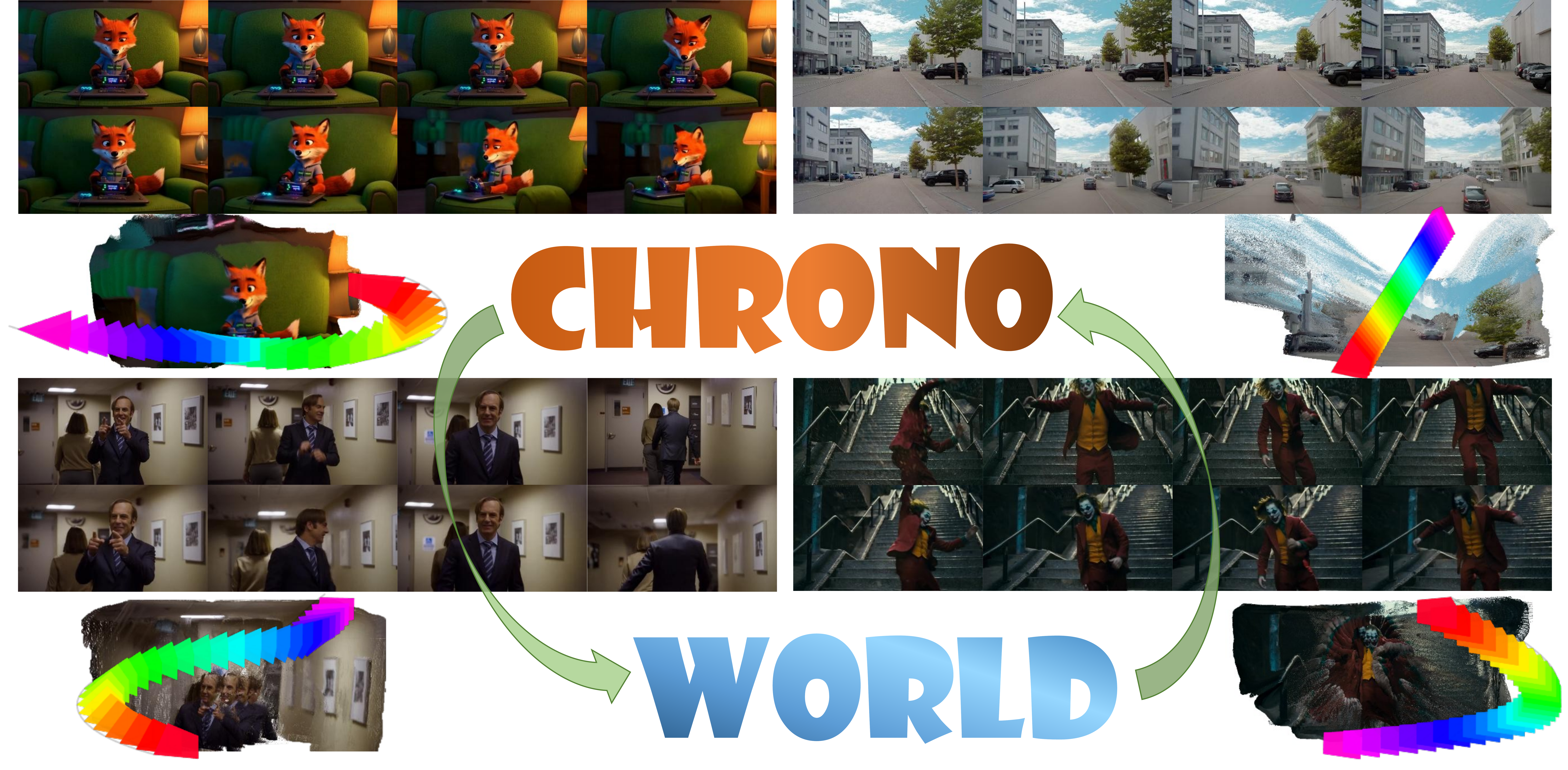}
\vspace{-10pt}
\caption{\emph{\textbf{Showcases of our \ourmethod{}.}} Our method addresses the challenge of spatiotemporally consistent 4D scene generation by introducing a ``Observation--State--Reflection'' framework. This approach leverages spatiotemporal causal cues and reconstruction-driven geometric reflection to generate globally consistent, free-view 4D world.
}
\label{fig:teaser}
\end{figure}
}

\begin{abstract}
While existing camera-controllable video generation models can produce visually compelling sequences, preserving intrinsic 4D spatiotemporal coherence remains challenging. To address this limitation, we propose \textbf{\ourmethod{}}, an ``Observation--State--Reflection'' framework that leverages spatiotemporal causal cues and reconstruction priors to generate globally consistent, free-view 4D scenes. Given a context video, we introduce a Spatiotemporal Epipolar Causal Attention mechanism that enforces multi-view epipolar constraints and temporal causality throughout the generation process. In addition, we develop a reconstruction-driven geometric reflection pipeline with a 4D retrieval strategy to enable dynamic self-assessment and correction of generated outputs, improving consistency and accuracy. Extensive experiments show that \ourmethod{} achieves state-of-the-art performance in spatiotemporally consistent, cinematic-quality 4D scene generation, with strong generalization and high-fidelity geometry across diverse scenarios.
\end{abstract}
\section{Introduction}
\label{intro}

Transforming sparse observations into freely navigable 4D worlds has long been a central goal of the field. Driven by recent advances in video generation and world models, modern generative architectures can synthesize cinematic-quality sequences and enable real-time interactive rendering. Beyond enabling cost-effective construction of immersive VR/AR environments~\cite{li2024advances, pages2018affordable}, these capabilities provide scalable, high-fidelity simulations that are critical for embodied intelligence~\cite{yang2024holodeck, wang2025embodiedgen} and autonomous driving~\cite{gao2024magicdrive3d, lu2025infinicube, wu2024holodrive}.

Despite these advances, a ``Tower of Babel'' gap persists between 2D video generation and 4D scene reconstruction. While current video generation models~\cite{wang2024motionctrl, he2024cameractrl, li2025realcam, geng2025motion} can produce visually plausible, pixel-level sequences, they struggle to maintain intrinsic 4D spatiotemporal consistency in the underlying scene representation. As a result, generated videos often violate multi-view geometric constraints, undermining both physical validity and temporal coherence.
Recent approaches~\cite{yu2025wonderworld, liu2025free4d, song2025worldforge, szymanowicz2025bolt3d} attempt to mitigate this issue by incorporating projected 3D representations (e.g., point clouds) as multi-view guidance within iterative render--generate--update pipelines. However, constructing 3D representations from coarse reconstructions of generated outputs introduces a critical failure mode: errors in synthesized frames are back-projected and fused into the 3D prior, creating a self-reinforcing feedback loop that amplifies inconsistencies and can ultimately lead to collapse in subsequent generations.

Fundamentally, these challenges can be attributed to three key limitations: \textbf{1) 4D Spatiotemporal Inconsistency:} Current models generate videos that violate 3D epipolar geometry and 4D spatiotemporal constraints, preventing the formation of a consistent and persistent scene representation. \textbf{2) Error Accumulation and Exposure Bias:} Due to training--inference discrepancies and inherited artifacts, errors at inference time accumulate over time, leading to pronounced dynamic distortions and spatial drift. \textbf{3) Spatial Forgetting and Geometric Degradation:} Over temporal horizons and complex camera motions, existing methods fail to preserve global coherence and handle occlusions, resulting in ``spatial forgetting'' that can cause severe geometric degradation or collapse.

To address these challenges, we draw inspiration from human cognition and propose an ``Observation--State--Reflection'' framework, \ourmethod{}, for camera-conditioned, globally consistent 4D scene generation. The framework extracts spatiotemporal cues from external inputs (\textbf{observations}), compresses them into a structured latent space via geometric invariants (\textbf{state}), and refines predictions through geometry-aware feedback (\textbf{reflection}), yielding a persistent and unified representation.
At its core, we introduce a Spatiotemporal Epipolar Causal Attention mechanism that leverages contextual views to preserve intra- and inter-frame dependencies, while enforcing multi-view epipolar constraints and temporal causality. We further propose a Spatial Context Augmentation strategy integrated with rectified flow, where perturbed memory cues encourage robustness and reduce error accumulation and exposure bias during spatial transformations. In addition, we augment the video diffusion model with a reconstructive multi-head decoder that maps latent features to 4D Gaussian attributes and motion fields, supported by auxiliary representation supervision to enhance spatiotemporal consistency.
To mitigate cumulative drift and address spatial forgetting and geometric degradation, we design a reconstruction-driven geometric reflection pipeline at inference time. Specifically, we build a unified 4D memory from feed-forward 4D Gaussians and global camera poses. An explicit 4D retrieval mechanism then evaluates generated outputs, filters inconsistent frames, and constructs a geometric consistency graph. Finally, we synchronize updates between the explicit memory and the implicit history cache to maintain accuracy and consistency over time.
The main contributions of our work are summarized as follows:
\begin{itemize}[nosep, wide, leftmargin=*]
\item We propose \textbf{\ourmethod{}}, an ``Observation--State--Reflection'' framework that bridges video generation and 4D reconstruction, enabling the synthesis of spatiotemporally consistent, free-view 4D scenes from input videos.
\item We introduce a Spatiotemporal Epipolar Causal Attention mechanism that exploits spatiotemporal cues to enforce multi-view epipolar constraints and temporal causality during generation.
\item We develop a reconstruction-driven geometric reflection pipeline with a 4D retrieval strategy to enable dynamic self-assessment and correction of generated outputs, improving accuracy and consistency of the 4D representation.
\item Extensive experiments show that \ourmethod{} achieves state-of-the-art performance in spatiotemporally consistent, cinematic-quality 4D scene generation, with strong generalization and high fidelity across diverse scenes and scales.
\end{itemize}

\section{Related Work}
\label{sec:related}

\noindent\textbf{Camera-Controlled Video Generation.}
Recent progress in video generation has been driven by large-scale diffusion models~\cite{yang2024cogvideox, ge2025arc, ma2025step, wan2025wan} and autoregressive transformer architectures~\cite{joseph2014nova, alhaija2025cosmos}. While these approaches achieve strong visual fidelity, conditioning on a single text or image input provides limited control over camera motion and restricts free-view exploration. Camera-conditioned methods~\cite{wang2024motionctrl, he2024cameractrl, zhang2025spatialcrafter} incorporate trajectory signals to guide video synthesis along prescribed camera paths. However, preserving geometric fidelity and scene consistency under large viewpoint changes remains challenging. Warping-and-repainting approaches project images into point clouds and warp them along trajectories, which improves spatial alignment but often introduces artifacts in unseen regions. Joint RGB-depth generation enables point cloud reconstruction for novel-view rendering, yet several limitations persist in 4D inconsistency.

To address these issues, we integrate diffusion-based video generation with both implicit and explicit geometric priors. In addition, we introduce a spatiotemporal self-reflection mechanism that enables the model to preserve consistency throughout the generation process.

\noindent\textbf{3D and 4D Scene Generation.}
Existing 3D/4D generation methods can be broadly categorized into: (1) compositional scene generation~\cite{zhou2024gala3d, po2024compositional, gao2024graphdreamer}, which assembles scenes from discrete 3D assets using layouts or scene graphs, and (2) end-to-end volumetric generation, which directly models scenes using representations such as occupancy fields, point clouds, voxels, or meshes. Compositional approaches improve object diversity and controllability but often struggle with realistic backgrounds and object--environment interactions. In contrast, end-to-end methods better preserve global semantic consistency, yet are constrained by limited 3D data, high annotation costs, and lower visual fidelity, and they do not fully benefit from large-scale 2D pretraining.

Our approach bridges this gap by integrating video generation with 4D reconstruction through spatiotemporal self-reflection and explicit 4D representation prediction, unifying generative modeling with geometric reasoning to produce spatially coherent 4D scenes.

\noindent\textbf{Generalizable Reconstruction.}
Traditional optimization-based approaches~\cite{mildenhall2021nerf, kerbl20233d, yu2024mip, zhang2024gaussian} rely on dense multi-view observations and per-scene optimization, which limits scalability. Feed-forward methods~\cite{wang2025vggt, jiang2025anysplat, keetha2025mapanything} incorporate geometric priors to handle sparse inputs, but they often fail to recover occluded or unseen regions.
Recent work combines generative models with reconstruction to improve multi-view consistency; however, performance remains dependent on accurate volumetric consistency, which are frequently violated in practice. In contrast, our method tightly couples controllable video generation with reconstruction-driven cue guidance, improving both geometric and semantic consistency and enabling photorealistic 4D reconstructive generation.

\section{Method}
\label{sec:method}

%%%%%%%%%%%%%%%%%%%%%%%%%%%%%%%%%%%%%%%

\begin{figure*}[t] 
\centering
\includegraphics[width=1.00\textwidth]{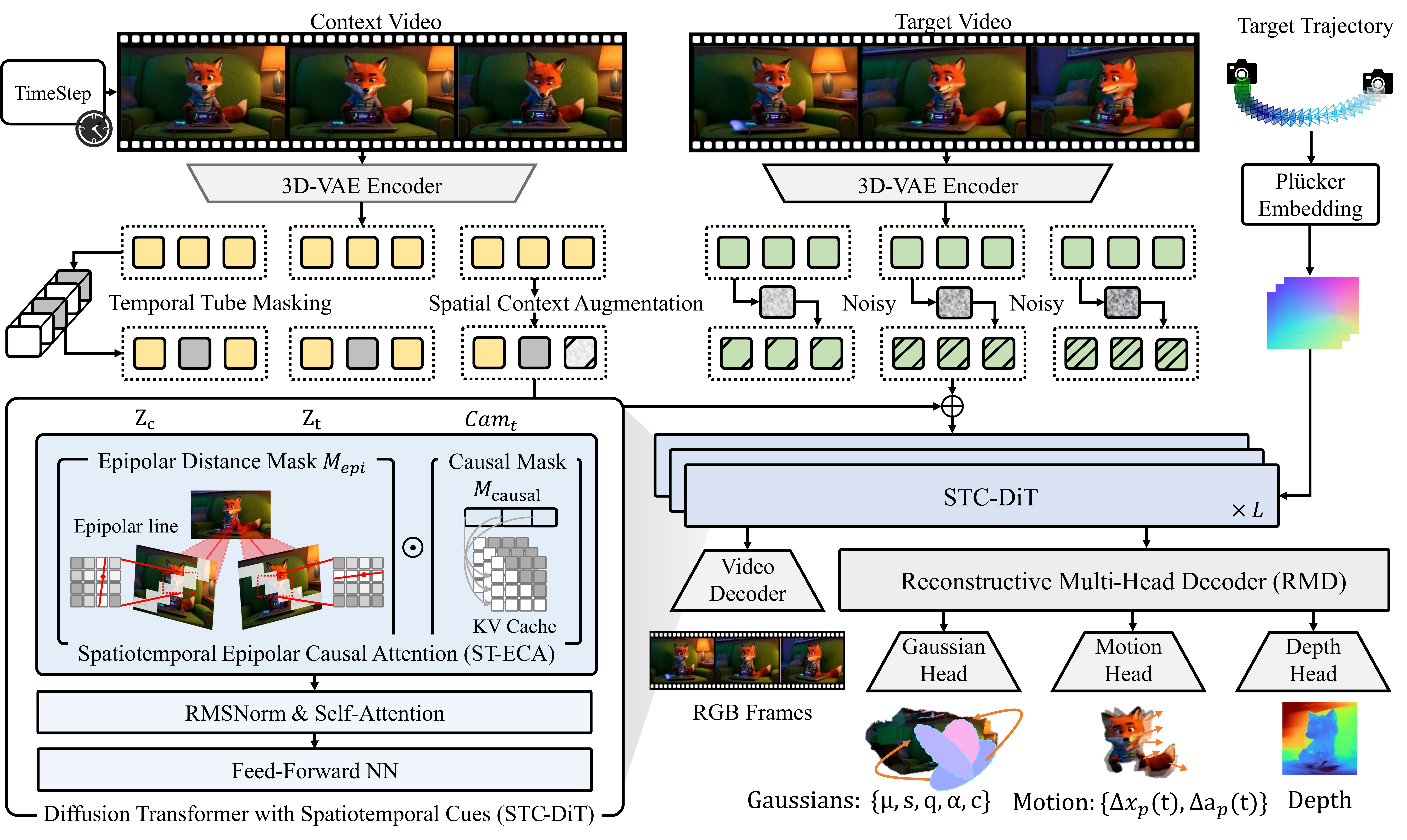}
\vspace{-20pt}
\caption{\textbf{Framework of \ourmethod{}.} Our method is trained to denoise target video conditioned on the camera trajectories with a Diffusion Transformer with Spatiotemporal Cues (STC-DiT), and a reconstructive multi-head decoder for predicting 4D attributes.}
\label{fig:overview}
\vspace{-10pt}
\end{figure*}

%%%%%%%%%%%%%%%%%%%%%%%%%%%%%%%%%%%%%%%

\begin{figure*}[t] 
\centering
\includegraphics[width=1.00\textwidth]{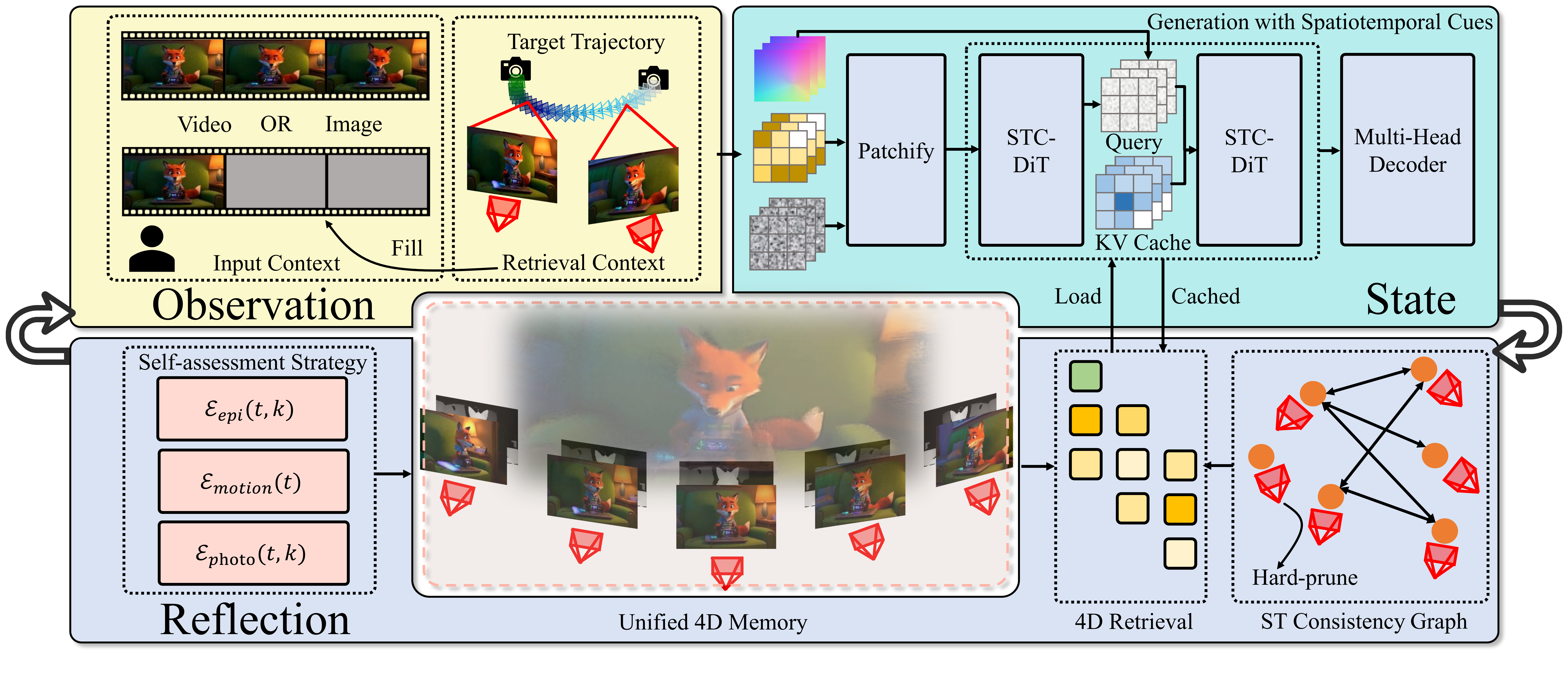}
\vspace{-20pt}
\caption{\textbf{``Observation-State-Reflection'' inference strategy.} Leveraging a unified 4D memory, \ourmethod{} enables dynamic self-assessment and correction of generated outputs, thereby enhancing the spatiotemporal consistency of 4D scenes.}
\label{fig:infer}
\vspace{-10pt}
\end{figure*}

This section presents \ourmethod{} (Fig.~\ref{fig:overview}-\ref{fig:infer}), which unifies video generation and reconstruction via a hybrid training--inference paradigm under an ``Observation--State--Reflection'' framework. During training, spatiotemporal cues and a reconstructive multi-head decoder enforce geometric consistency in the latent space. During inference, a reconstruction-driven geometric reflection mechanism refines outputs and maintains a consistent 4D memory.

\subsection{Overview Architecture}

Our architecture builds on a latent video diffusion framework, comprising a causal 3D VAE encoder, a Diffusion Transformer with Spatiotemporal Cues (STC-DiT), and a reconstructive multi-head decoder (Fig.~\ref{fig:overview}). During training, we sample triplets of context videos, target videos, and target camera trajectories, where context and target sequences are temporally aligned but captured from different viewpoints. The causal 3D VAE encodes the inputs into latent representations $Z_c, Z_t \in \mathbb{R}^{B \times C \times T \times H \times W}$, where $B$, $C$, $T$, $H$, and $W$ denote batch size, channels, timesteps, height, and width, respectively. We encode the camera trajectory $\{(T_i, R_i)\}_{i=0}^{T-1}$ into dense positional features using Plücker embeddings, followed by a two-layer MLP modulation network that injects the features into the attention keys $K$ and values $V$. 

To simulate inference-time degradation, we apply Temporal Tube Masking and Spatial Context Augmentation to the context latents $Z_c$, modeling error accumulation and information loss. The model is trained to denoise $Z_t$ conditioned on $Z_c$ and the camera trajectory, enabling temporally coherent generation and globally consistent 4D geometry across the sequence. Finally, the multi-head decoder converts the block-wise, asynchronously denoised latents into explicit outputs, including RGB videos (appearance), Gaussian attributes (geometry), motion maps (dynamics), and depth maps.

\subsection{Diffusion Transformer with Spatiotemporal Cues}

Existing diffusion models typically capture intra-frame spatial and inter-frame temporal dependencies via full-sequence cross- or causal attention, largely overlooking cross-frame spatiotemporal cues (i.e., latent correspondences along epipolar trajectories). This oversight introduces computational redundancy and degrades geometric consistency across overlapping regions.

To address this issue, we propose a diffusion framework with Spatiotemporal Epipolar Causal Attention. By coupling multi-view geometric constraints with spatiotemporal augmentation, the model enforces coherent cross-view correspondence, enabling the generation of consistent 4D scenes.

\noindent \textbf{Spatiotemporal Epipolar Causal Attention.}
To reduce the computational redundancy and geometric ambiguity of unconstrained full-sequence attention, we propose spatiotemporal epipolar causal attention (ST-ECA). By integrating multi-view 3D priors with strict temporal causality, ST-ECA confines cross-frame interactions to geometrically valid, causally ordered regions.

Given query $\mathbf{Q} \in \mathbb{R}^{N \times d}$, key $\mathbf{K} \in \mathbb{R}^{L \times d}$, and value $\mathbf{V} \in \mathbb{R}^{L \times d}$ projected from latent features, we first construct an epipolar distance mask $\mathbf{M}_{\text{epi}}$ to enforce multi-view geometric consistency at timestep $T$. For a query token at 2D coordinate $\mathbf{p}_i$ in the target view and a key token at $\mathbf{p}_j$ in the context view, valid correspondences lie on the epipolar line $\mathbf{l}_i = \mathbf{F} \mathbf{p}_i$, where $\mathbf{F}$ is the fundamental matrix induced by relative camera poses. We define the epipolar mask as:
\begin{equation}
    \mathbf{M}_{\text{epi}}(i, j) = -\gamma \frac{|\mathbf{p}_j^T \mathbf{F} \mathbf{p}_i|}{\|\mathbf{F} \mathbf{p}_i\|_2},
\end{equation}
where $\gamma$ is a learnable scaling factor controlling the sharpness of the geometric constraint.

To enforce temporal causality and prevent leakage from future states, we introduce a binary temporal mask $\mathbf{M}^{\text{causal}}_{i,j} \in \{0, -\infty\}$, defined as $0$ if $\tau(i) \ge \tau(j)$ and $-\infty$ otherwise, where $\tau(\cdot)$ denotes the frame timestamp. Each token can thus attend only to the current and past frames.

We integrate both constraints by injecting the geometric and causal biases into the attention:
\begin{equation}
    \text{ST-ECA}(\mathbf{q}, \mathbf{k}, \mathbf{v}) = \sum_{j} \text{Softmax}\left( \frac{\mathbf{q}_i \mathbf{k}_j^T}{\sqrt{d}} + \mathbf{M}_{\text{epi}}(i,j) + \mathbf{M}_{\text{causal}}(i,j) \right) \mathbf{v}_j,
\end{equation}
where $d$ is the feature dimension. The epipolar mask acts as a continuous geometric prior, restricting correspondence search to 1D epipolar trajectories rather than the full 2D plane, while the causal mask preserves temporal consistency and reduces the cost of global attention.

\noindent \textbf{Temporal Tube Masking.} 
To improve temporal reasoning and motion prediction, naive approaches apply independent spatial masks per frame, which allows trivial reconstruction by copying from adjacent frames. To prevent this shortcut, we introduce Temporal Tube Masking (TTM), which occludes consistent spatial regions across the entire temporal sequence.

We first sample a 2D binary spatial mask $\mathbf{M}_s \in \{0, 1\}^{H \times W}$ from a Bernoulli distribution, $\mathbf{M}_s(x, y) \sim \text{Bernoulli}(p_{\text{drop}})$, where $(x, y)$ denotes coordinates and $p_{\text{drop}}$ is the masking ratio. This mask is then extended along the temporal dimension to form a 4D tube mask $\mathbf{M}_{\text{tube}}(t, x, y) = \mathbf{M}_s(x, y)$ for all $t \in \{1, \dots, T\}$. The corrupted context latent is obtained via element-wise masking:
\begin{equation}
    \hat{\mathbf{Z}}_c = \mathbf{Z}_c \odot (\mathbf{1} - \mathbf{M}_{\text{tube}}),
\end{equation}
where $\odot$ denotes the Hadamard product.
By enforcing consistent occlusions over time, TTM prevents degenerate temporal copy-paste behavior and instead encourages reconstruction from long-range geometric cues and motion dynamics. 

\noindent \textbf{Spatial Context Augmentation.}
To mitigate exposure bias and spatial drift (error accumulation) in video generation, we introduce Spatial Context Augmentation (SCA), a lightweight self-augmentation strategy in the flow-matching latent space. Given ground-truth context and target frames, the causal 3D VAE encodes them into clean latents $\mathbf{Z}_c^0$ and $\mathbf{Z}_t^0$. To simulate inference-time degradation, we corrupt the context latent by sampling a flow timestep $\tau \sim \mathcal{U}(0, 0.5)$ and injecting noise via the rectified flow schedule:
\begin{equation}
    \mathbf{Z}_c^\tau = (1 - \tau) \mathbf{Z}_c^0 + \tau \epsilon,
\end{equation}
where $\epsilon \sim \mathcal{N}(0, \mathbf{I})$ and $\tau$ controls the corruption level. To avoid training collapse, $\tau$ is constrained to be lower than the noise level applied to the target latent.

The flow matching objective then supervises recovery of the target $\mathbf{Z}_t^0$ from corrupted context. This structured perturbation improves robustness to cumulative drift by explicitly training the model to correct errors induced by degraded spatial context.

\noindent \textbf{Efficient Training Strategy.}
To reduce the computational cost and stabilize optimization in video diffusion, we introduce an efficient training strategy that leverages spatiotemporal structure for accelerated generation. We assign monotonically increasing noise levels $\tau_i$ along the temporal axis to enforce causal progression.

We then partition spatial latents into non-overlapping blocks $\{\mathcal{B}_k\}$ and perform block-wise asynchronous denoising with locally cached context:
\begin{equation}
    \mathbf{Z}_{\tau-\Delta \tau}^{\mathcal{B}_k} = \mathbf{Z}_\tau^{\mathcal{B}_k} - \Delta \tau \cdot v_\theta(\mathbf{Z}_\tau^{\mathcal{B}_k}, \tau, \mathbf{C}_{\neg k}),
\end{equation}
where $\mathbf{C}_{\neg k}$ denotes cached key-value features from neighboring blocks and $v_\theta(\cdot)$ is the velocity field. This design reduces redundant computation and decouples gradients across blocks.

Finally, we combine this scheme with rectified flow for few-step generation. We apply velocity-matching distillation, where the student $v_\theta$ learns a target velocity pointing to the teacher's multi-step prediction, reducing the denoising process from $\sim$50 steps to 5 steps.

\subsection{Reconstructive Multi-Head Decoder}

Directly mapping high-dimensional denoised latents to dynamic scenes is highly under-constrained. We address this by employing a dual-stage training with four parallel decoders that disentangle the output space into complementary attributes: RGB frames (appearance), Gaussian attributes (geometry), motion maps (dynamics), and depth maps.

\noindent \textbf{Appearance decoder.}
We use the pre-trained 3D VAE decoder to project latents back to pixel space, producing $\mathbf{I} \in \mathbb{R}^{T \times 3 \times H \times W}$. To preserve the learned latent manifold, this decoder is kept frozen while training the remaining heads.

\noindent \textbf{Geometry, motion, and depth decoders.}
We adopt a Dense Prediction Transformer (DPT) architecture~\cite{wang2025vggt, wang20254d}. Given denoised tokens $\mathbf{Z} \in \mathbb{R}^{N \times d}$, we reassemble them into multi-scale feature maps $\mathcal{F}_l \in \mathbb{R}^{H_l \times W_l \times C_l}$ across transformer layers $l$. A fusion module with residual convolutions and progressive upsampling aggregates these features into a dense representation $\mathbf{F}_{\text{dense}}$.
Following~\cite{wu2025geometry}, we apply geometry forcing to align intermediate diffusion features with geometry-aware supervision from 4D reconstruction models~\cite{lin2025movies}.

The Gaussian decoder maps $\mathbf{F}_{\text{dense}}$ to per-pixel 3D Gaussian attributes in canonical space. For each splat $g$, it predicts
$A_g = \{\mathbf{\mu}, \mathbf{s}, \mathbf{q}, \alpha, \mathbf{c}\}$,
where $\mathbf{\mu} \in \mathbb{R}^3$ is position, $\mathbf{s} \in \mathbb{R}^3$ anisotropic scale, $\mathbf{q} \in \mathbb{R}^4$ rotation quaternion, $\alpha \in [0,1]$ opacity, and $\mathbf{c}$ spherical harmonics coefficients.

To model dynamics, the motion decoder predicts per-pixel displacements $m_p(t) = \{\Delta \mathbf{x}_p(t), \Delta \mathbf{a}_p(t)\}$ at query time $t$, where $\Delta \mathbf{x}_p(t) \in \mathbb{R}^3$ is the 3D translation relative to the canonical position $\mathbf{\mu}_p$, and $\Delta \mathbf{a}_p(t)$ captures temporal variations in other attributes (e.g., rotation or scale). This decoupling of static geometry and dynamic motion enables explicit temporal tracking of each Gaussian.

An auxiliary depth decoder predicts a dense depth map $\mathbf{D} \in \mathbb{R}^{H \times W}$, where $d_p$ provides geometric regularization to improve spatial consistency and resolve scale ambiguity. Overall, the DPT-based design translates tokenized latents into pixel-aligned Gaussian primitives with coherent dynamics.

To supervise geometry, motion, and depth, we derive pseudo-labels from pre-trained reconstruction models~\cite{lin2025movies, lin2025depth}, caching their intermediate features as geometry-aware signals to align with diffusion hidden states via the reconstructive multi-head decoder.

%%%%%%%%%%%%%%%%%%%%%%%%%%%%%%%%%%%%%%%

%%%%%%%%%%%%%%%%%%%%%%%%%%%%%%%%%%%%%%%

\subsection{Reconstruction-driven Geometric Reflection}

 A key limitation of existing video and 4D scene generation methods is that they lack reliable mechanisms to assess the quality of their own predictions, violating spatiotemporal consistency under zero-shot conditions. While recent 4D memory-augmented and post-reconstruction approaches attempt to enhance coherence, their uncorrected fusion of flawed predictions exacerbates error accumulation, inducing semantic drift, spatial forgetting, and geometric inconsistency.

To address this, we propose an \emph{Observation-State-Reflection} inference strategy with Reconstruction-driven Geometric Reflection. The framework performs iterative retrieval, selection, and self-refinement, breaking the accumulation of single-pass errors and establishing a feedback loop for correction. This enables robust, globally consistent 4D scene generation across time.

\noindent \textbf{Unified 4D Memory Construction.}

While existing 3D-enhanced video generation methods passively unproject generated outputs into local spatial representations (e.g., depth maps, point clouds), their lack of global 4D modeling restricts them to static scenes and exacerbates error accumulation. To address this, we propose a unified 4D memory construction strategy that aligns and fuses predicted representations within a Canonical Gaussian Space.

Specifically, we fix the first context view as the global reference frame $\mathcal{C}_{\text{ref}}$, and transform target camera poses $\mathbf{T}_{\text{tgt}}$ into this frame via $\mathbf{T}_{\text{rel}} = \mathbf{T}_{\text{ref}}^{-1}\mathbf{T}_{\text{tgt}}$. This ensures all Gaussians are generated in a consistent canonical space, mitigating global pose drift. For each splat $p$, the dynamic 4D Gaussian state at time $t$ is constructed by augmenting the canonical base with predicted motion:
\begin{equation}
    \mathbf{x}_p(t) = \mu_p + \Delta \mathbf{x}_p(t), \quad \mathcal{A}_p(t) = \mathcal{A}_p \oplus \Delta \mathbf{a}_p(t),
\end{equation}
where $\mathbf{x}_p(t) \in \mathbb{R}^3$ is the Gaussian center at time $t$, $\mu_p \in \mathbb{R}^3$ is the canonical position anchored in $\mathcal{C}_{\text{ref}}$, and $\Delta \mathbf{x}_p(t)$ is the predicted motion offset. $\mathcal{A}_p(t)$ denotes time-varying attributes (e.g., rotation, scale), updated via $\Delta \mathbf{a}_p(t)$ using attribute-specific operators (e.g., quaternion composition).

The predicted depth map $\mathbf{D}$ further serves as a scale regularizer, grounding the 4D representation by normalizing Gaussian coordinates into a consistent metric range (e.g., $[-1,1]^3$).

\noindent \textbf{Self-Assessing for Reflection.}
To ensure fidelity of the unified 4D representation and suppress hallucinated structures, we introduce a hybrid self-assessment strategy that combines epipolar geometry, motion consistency, and dense reprojection errors. For a generated block at timestep $t$, the model constructs canonical dynamic Gaussians $\mathcal{G}_t$. We evaluate geometric reliability by differentiably reprojecting (rasterizing) $\mathcal{G}_t$ onto camera poses $\mathbf{T}_k$ of a retrieved set of verified anchor frames $\mathcal{M}_{\text{ref}}$.

We first compute the epipolar error using the symmetric Sampson distance:
\begin{equation}
    \mathcal{E}_{\text{epi}}(t, k) =
    \frac{1}{\|\mathbf{M}_{t \to k}\|_0}
    \sum_{\mathbf{u}_t \leftrightarrow \mathbf{u}_k}
    \mathbf{M}_{t \to k}(\mathbf{u}_t)
    \frac{(\mathbf{u}_t^T \mathbf{F}_{t \to k} \mathbf{u}_k)^2}
    {\| \mathbf{F}_{t \to k} \mathbf{u}_k \|_2^2 + \| \mathbf{F}_{t \to k}^T \mathbf{u}_t \|_2^2},
\end{equation}
where $\mathbf{u}_t, \mathbf{u}_k \in \mathbb{R}^3$ are corresponding homogeneous pixels, $\mathbf{F}_{t \to k}$ is the fundamental matrix between $\mathbf{T}_t$ and $\mathbf{T}_k$, and $\mathbf{M}_{t \to k}$ is a depth-aware co-visibility mask.
We then measure motion consistency via:
\begin{equation}
    \mathcal{E}_{\text{motion}}(t) =
    \mathbb{E}_{p,q}\!\left[\|\Delta \mathbf{x}_p - \Delta \mathbf{x}_q\|_2^2\right]
    + \lambda \mathbb{E}_{p}\!\left[\|\partial_t \Delta \mathbf{a}_p\|_2^2\right],
\end{equation}
where $\Delta \mathbf{x}_p$ and $\Delta \mathbf{x}_q$ denote motion vectors of temporally adjacent Gaussians, and $\partial_t \Delta \mathbf{a}_p$ penalizes high-frequency attribute variation.
Finally, we evaluate photometric and perceptual consistency by comparing rendered projections $\hat{\mathbf{I}}_{t \to k}$ with observations $\mathbf{I}_k$:
\begin{equation}
    \mathcal{E}_{\text{photo}}(t, k) =
    \|\mathbf{M} \odot (\hat{\mathbf{I}}_{t \to k} - \mathbf{I}_k)\|_1
    + \lambda \mathcal{L}_{\text{LPIPS}},
\end{equation}
where $\mathbf{M}$ is a co-visibility mask and $\mathcal{L}_{\text{LPIPS}}$ is the perceptual loss~\cite{dosovitskiy2016generating}.
These metrics are fused into a scale-invariant consistency graph spanning the current frame and historical context, enabling robust identification of unreliable predictions.

\noindent \textbf{Cycle Refinement via 4D Retrieval Selection.}
Given an observation $\mathcal{O}$, our STC-DiT and reconstructive multi-head decoder output complementary representations (RGB, Gaussian parameters, motions, and depth) to construct a Unified 4D Memory. To prevent flawed frames from corrupting generations via passive caching, we formulate this memory as a dynamically evaluated topological graph. At inference, we execute a cyclic refinement loop: the model self-assesses this graph, hard-prunes geometrically inconsistent nodes, and retrieves the top-$K$ reliable frames to dynamically refresh the KV cache. Conditioned on this curated context, corrupted regions are iteratively regenerated, thereby halting error propagation and reducing computational costs.

\noindent \textbf{Active 4D Retrieval and KV Cache Refresh.}
Unlike conventional memory-augmented video generation that retrieves historical RGB frames or latent tokens based on appearance similarity, temporal proximity, or view overlap, our method retrieves structured 4D states in a unified Canonical Gaussian Space. Each state jointly represents persistent geometry, time-varying attributes, explicit motion, and depth. For the $p$-th Gaussian primitive, its position at time $t$ is represented as $\mathbf{x}{p}(t)=\boldsymbol{\mu}{p}+\Delta\mathbf{x}{p}(t)$, where $\boldsymbol{\mu}{p}$ is the canonical position and $\Delta\mathbf{x}{p}(t)$ is its temporal motion. Its time-varying attributes are similarly defined as $\mathcal{A}{p}(t)=\mathcal{A}{p}\oplus\Delta\mathbf{a}{p}(t)$. This formulation explicitly associates the same Gaussian primitive across time, turning historical observations into a queryable 4D memory rather than independent frame-level context.

At each inference step, we first assess the reliability of historical states using the reconstruction-driven consistency graph, which jointly considers epipolar geometry, motion consistency, and reprojection fidelity. Denoting the aggregated inconsistency of memory node $m_i$ by $\Phi_i$, we retain only states satisfying $\Phi_i\leq\tau_{\mathrm{prune}}$, forming the clean memory $\mathcal{M}{t}^{\mathrm{clean}}$. This hard pruning is performed before retrieval, preventing geometrically inconsistent states from contaminating subsequent generation. Given the current 4D state as query $q_t$, we then perform active Top-$K$ retrieval according to the joint spatiotemporal ranking function $S{\mathrm{ST}}(m_i,q_t)$, yielding $\mathcal{H}t=\operatorname{TopK}{m_i\in\mathcal{M}{t}^{\mathrm{clean}}}S{\mathrm{ST}}(m_i,q_t)$. Here, relevance is determined by the joint geometric validity and spatiotemporal utility of each state rather than appearance similarity alone, allowing reliable long-range observations to be actively recalled when they provide useful constraints for the current generation.

The retrieved states are encoded into the generator's key-value context as $(\mathbf{K}_t,\mathbf{V}t)=\operatorname{Encode}{\mathrm{KV}}(\mathcal{H}_t)$, which replaces the previously accumulated historical context. Thus, the active KV cache is bounded by the retrieval budget $K$, rather than the total history length, reducing the context size from $O(N_tL)$ to $O(KL)$ when each state contributes $L$ tokens. Importantly, this is selective compression rather than temporal truncation, since states are retained according to both reliability and query relevance.

The retrieval is performed recurrently throughout inference. Each generation step produces new RGB, Gaussian geometry, motion, and depth, which extend the Unified 4D Memory and are subsequently re-evaluated to update the consistency graph. The refined memory then undergoes hard pruning, Top-$K$ retrieval, and KV refresh for the next generation cycle, forming a closed-loop ``generate $\rightarrow$ assess $\rightarrow$ prune $\rightarrow$ retrieve $\rightarrow$ refresh $\rightarrow$ regenerate'' process. In this way, 4D RAG preserves scene information by actively recalling reliable states beyond the current temporal window, maintains a bounded inference context by restricting the active KV cache to $K$ retrieved states, and suppresses cumulative spatial and temporal drift by preventing inconsistent predictions from being repeatedly propagated through memory. This active, reliability-aware memory update is therefore essential for maintaining stable 4D reconstruction and generation over extended sequences.

\section{Experiments}
\label{sec:exp}

\subsection{Experimental Settings}

\begin{table*}[t]
\small
\footnotesize
\centering
\caption{\textbf{Quantitative evaluation} of our method against state-of-the-art models in camera-controlled video generation, 3D-enhanced scene generation, 4D dynamic scene generation, measuring visual generation quality, and 4D visual synchronization. All tests are conducted under fair comparison settings, using identical video inputs and target camera trajectories.}
\resizebox{1.0\linewidth}{!}{
\begin{tabular}{llccccccccc}
    \toprule
    \multirow{2}{*}{Method} & \multirow{2}{*}{Type} & \multicolumn{2}{c}{Visual Quality}    & \multicolumn{3}{c}{4D Visual Synchronization}  & \multicolumn{2}{c}{Camera Accuracy} \\ \cmidrule(rl){3-4} \cmidrule(l){5-7} \cmidrule(l){8-9}
    & & CLIP-V$\uparrow$ & FID$\downarrow$ & FVD-F$\downarrow$ & FVD-V$\downarrow$ & FVD-4D$\downarrow$ & RPE-R$\downarrow$ & RPE-T$\downarrow$\\ 
    \midrule
    \small{MotionCtrl~\cite{wang2024motionctrl}} & 2D & 0.46 & 123.68 & 535.96 & 421.58 & 442.72 & 4.77 & 8.93 \\
    \small{CameraCtrl~\cite{he2024cameractrl}} & 2D & 0.49 & 114.27 & 510.59 & 378.05 & 390.16 & 4.02 & 7.65 \\
    \small{RecamMaster~\cite{bai2025recammaster}} & 2D & 0.64 & 101.58 & 403.18 &  214.33 &  198.52 & 3.01 & 6.24 \\
    \midrule
    \small{TrajectoryCrafter~\cite{yu2025trajectorycrafter}} & 3D & 0.64 & 98.46 & 487.02 & 293.56 & 286.50 & 1.26 & 3.76 \\
    \small{Vmem~\cite{li2025vmem}} & 3D & 0.66 & 113.41 & 389.64 & 252.08 & 264.73 & 2.59 & 6.42 \\
    \small{GEN3C~\cite{ren2025gen3c}}  & 3D & 0.58 & 87.60 & 474.86 & 274.71 & 283.48 & 2.76 & 5.21 \\
    \midrule
    \small{DeepVerse~\cite{chen2025deepverse}} & 4D & 0.67 & 72.51 & 376.05 & 237.10 & 239.77 & 1.82 & 3.20 \\
    \small{Free4D~\cite{liu2025free4d}} & 4D & 0.71 & 88.32 & 347.56 & 224.95 & 216.67 & 1.73 & 3.28 \\
    \small{Neoverse~\cite{yang2026neoverse}} & 4D & 0.72 & 90.16 & 353.48 & 246.71 & 223.10 & 1.68 & 3.02 \\
    \midrule
    \textbf{\small{Ours}} & 4D & \textbf{0.80} & \textbf{64.58} & \textbf{314.28} &  \textbf{201.46} &  \textbf{177.41} &  \textbf{0.95} & \textbf{2.34} \\
    \bottomrule
\end{tabular}
\label{visual-compare}
}
\end{table*}
\vspace{-10pt}

\noindent \textbf{Implementation Details.}
Our framework builds on Wan2.1 I2V~\cite{wan2025wan}, augmented with a Diffusion Transformer with Spatiotemporal Cues (STC-DiT), Reconstructive Multi-Head Decoder (RMD), and camera pose embeddings with spatial context augmentation. We train on a mixture of public video generation and reconstruction datasets~\cite{bai2025recammaster, ling2024dl3dv}, combining real and synthetic dynamic scenes. % To balance sources, we uniformly sample 10K videos from each dataset. Camera parameters are estimated using MoVieS~\cite{lin2025movies}. 
We perform rectified flow distillation for STC-DiT with learning rate $1\times10^{-3}$, batch size 10, and 20K steps, using AdamW (weight decay 0.05) and a cosine schedule with linear warm-up. Subsequently, the RMD is trained incorporating geometric forcing, with a learning rate of $1\times10^{-6}$ and a batch size of 10. DPM++ is used for efficient epipolar sampling. Training runs on 8$\times$A100 (80GB) GPUs, while inference uses a single A100 GPU. Additional details are provided in the Supplementary Material.

\noindent \textbf{Evaluation.}
We curate a comprehensive testing set of real and synthetic videos, which are sourced from diverse datasets, encompassing movies, games, animations, real-world captures, indoor scenes, in-the-wild driving scenarios, aerial drone footage, and web-sourced videos. We assess the model using multi-dimensional metrics: visual quality (CLIP-V, FID), 4D visual fidelity (FVD-F, FVD-V, FVD-4D), camera pose accuracy (RPE-R, RPE-T), reconstruction quality (PSNR, SSIM, LPIPS). VBench~\cite{huang2024vbench, zheng2025vbench} is also employed to measure spatiotemporal consistency.

%%%%%%%%%%%%%%%%%%%%%%%%%%%%%%%%%%%%%%%
\begin{figure*}[t] 
\centering
\includegraphics[width=1.00\textwidth]{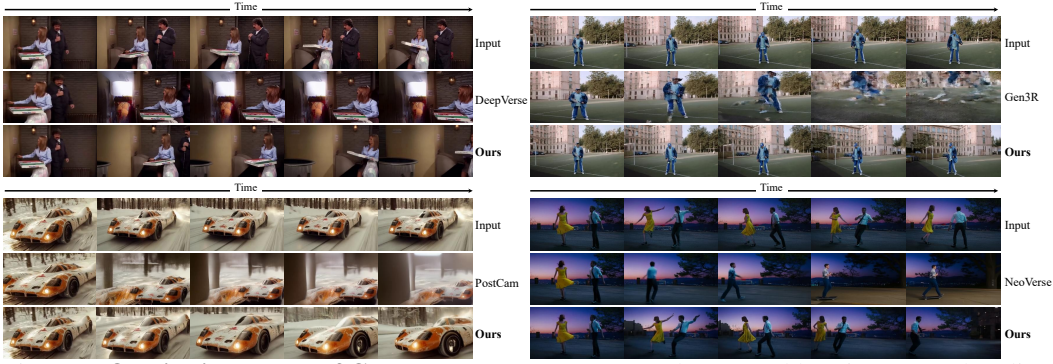}
\vspace{-20pt}
\caption{\textbf{Qualitative results of \ourmethod{}.} We compare \ourmethod{} against SOTA baselines across diverse camera trajectories to validate its effectiveness (zoom in for details).}
\label{fig:vis1}
\vspace{-10pt}
\end{figure*}
%%%%%%%%%%%%%%%%%%%%%%%%%%%%%%%%%%%%%%%

\subsection{Comparisons with State-of-the-Art Methods}
We compare our proposed~\ourmethod{} with state-of-the-art approaches in camera-controlled video generation, 3D-enhanced scene generation, and 4D dynamic scene generation.

\noindent \textbf{Quantitative Comparison.}
As shown in Tab.~\ref{visual-compare}, our method consistently surpasses all baselines across all metrics, achieving stronger visual quality alongside significantly improved 4D consistency and trajectory accuracy. Among the baselines, camera-controlled video generation methods~\cite{bai2025recammaster, wang2024motionctrl, he2024cameractrl, zhou2025stable} generate visually plausible videos but suffer from shape drift and trajectory misalignment without 4D constraints and explicit reflection. Current 3D-enhanced scene generation methods~\cite{li2025vmem, yu2024viewcrafter} rely on static surfel- or point-based memory, struggling to synthesize high-fidelity, temporally coherent 4D dynamic scenes. Despite utilizing depth-warped conditioning, GEN3C~\cite{ren2025gen3c} and TrajectoryCrafter~\cite{yu2025trajectorycrafter} introduce pronounced textural artifacts stemming from unnatural geometric priors and flawed projections.

While existing 4D dynamic scene generation methods~\cite{liu2025free4d, chen2025deepverse, bahmani2025lyra, yao2025sv4d, yang2026neoverse} utilize 4D memory, their naive ``generate-and-reconstruct'' relies on expensive post-reconstruction and per-frame querying, inevitably amplifying error accumulation and inconsistencies. In contrast, our model achieves globally consistent conditional generation by aligning the implicit latent and 4D spaces via spatiotemporal constraints, and rectifying discrepancies through explicit geometric reflection.

\noindent \textbf{Qualitative Comparison.}
As shown in Fig.~\ref{fig:vis1}, we visualize the 4D scenes generated from a source video along arbitrary camera trajectories. Compared with baselines, our approach generates high-fidelity, spatiotemporally coherent 4D scenes, exhibiting realistic appearance, precise pose accuracy, and stable geometric structures. More results and videos are provided in the Supplementary Material.

\begin{table*}[t]
\small
\footnotesize
\centering
\caption{\textbf{Quantitative comparisons on spatiotemporal consistency} of our method against state-of-the-art models in generation and reconstruction of 4D scenes.}
\resizebox{1.0\linewidth}{!}{
\begin{tabular}{llccccccccc}
    \toprule
    \multirow{2}{*}{Method} & \multicolumn{3}{c}{Representation Consistency}    & \multicolumn{6}{c}{Spatiotemporal Consistency}   \\ \cmidrule(rl){2-4} \cmidrule(l){5-10}
    & PSNR$\uparrow$  & SSIM$\uparrow$ & LPIPS$\downarrow$ & \makecell[c]{Aesthetic\\Quality $\uparrow$} & \makecell[c]{Imaging\\Quality $\uparrow$} & \makecell[c]{Temporal\\Flickering $\uparrow$} & \makecell[c]{Motion\\Smoothness $\uparrow$} & \makecell[c]{Subject\\Consistency $\uparrow$} & \makecell[c]{Background\\Consistency $\uparrow$} \\ 
    \midrule
    \small{Free4D~\cite{liu2025free4d}} & 18.03 & 0.64 & 0.48 & 0.51 & 0.54 & 0.96 & 0.98 & 0.92 & 0.94\\
    \small{More4D~\cite{zhang2025joint}}  & 18.32 & 0.67 & 0.49 & 0.48 & 0.56 & 0.97 & 0.98 & 0.91 & 0.95 \\
    \small{DeepVerse~\cite{liu2025free4d}} & 16.37 & 0.56 & 0.61 & 0.50 & 0.51 & 0.92 & 0.95 & 0.93 & 0.91 \\
    \small{Lyra~\cite{bahmani2025lyra}}  & 17.01 & 0.58 & 0.72 & 0.41 & 0.43 & 0.88 & 0.81 & 0.87 & 0.90 \\
    \small{Gen3R~\cite{huang2026gen3r}}  & 17.28 & 0.64 & 0.68 & 0.44 & 0.46 & 0.49 & 0.88 & 0.90 & 0.92 \\
    \midrule
    \textbf{\small{Ours}}  & \textbf{18.87} &  \textbf{0.73} & \textbf{0.45} & \textbf{0.52} & \textbf{0.58} & \textbf{0.99} & \textbf{0.99} & \textbf{0.95} & \textbf{0.96} \\
    \bottomrule
\end{tabular}
\label{geo-compare}
}
\end{table*}
\vspace{-5pt}

\subsection{Comparisons on Spatiotemporal Consistency}
In this work, we aim to enhance the spatiotemporal consistency of scene generation, jointly synthesizing cinematic-quality videos and geometrically coherent 4D representations. Thus, we validate the unified 4D memory by rendering target novel views and quantifying their photometric and spatiotemporal fidelity. Tab.~\ref{geo-compare} shows that our approach consistently outperforms our main competitors in preserving multi-view geometry and spatiotemporal consistency for 4D scenes. Notably, our Spatiotemporal Epipolar Causal Attention enforces latent geometric alignment, while Reconstruction-driven Geometric Reflection iteratively rectifies textural artifacts and spatial distortions.

\subsection{Ablation Studies}
We perform comprehensive ablation studies to validate the key components of our framework, as summarized in Tab.~\ref{ablation}. Further detailed ablations are deferred to the Supplementary Material.

Spatiotemporal Epipolar Causal Attention (ST-ECA) and Plücker ray embeddings jointly enforce multi-view geometric coherence and trajectory alignment. The introduced temporal tube masking, alongside spatial context augmentation, facilitates the mitigation of exposure bias and the rectification of spatial distortions. Reconstruction-driven Geometric Reflection proves effective in mitigating spatial forgetting and geometric degradation by enabling the generative model to iteratively rectify errors through self-reflection. Our proposed dual-stage training, coupled with geometry forcing, enables the model to decode 4D scene representations from latent features, thereby significantly enhancing the spatial consistency of predictions while accelerating the cyclic inference by 10$\times$.

\definecolor{best}{rgb}{0.996, 0.769, 0.3098}

\definecolor{second}{rgb}{1, 0.969, 0.733}

\begin{table*}[ht]
\small
\footnotesize
\centering
    \caption{\textbf{The effect of core components in our method.}}
    %\resizebox{1.0\linewidth}{!}{
    \scalebox{0.9}{
    \begin{tabular}{lcccc}
        \toprule
        {Method} & PSNR$\uparrow$ & CLIP-V$\uparrow$ & FVD$\downarrow$ & $RPE_{avg}$$\downarrow$ \\ 
        %\midrule
        \midrule
        \small{w/o ST-ECA}  & 18.03 & 0.71 & 184.27 & 1.75 \\
        \small{w/o Plücker Emb}  & 18.71 & 0.79 & 181.42 & 1.79  \\
        \small{w/o RMD}  & 18.59 & 0.77 & 212.79 & 1.68 \\
        \small{w/o ST Reflection}  & 17.65 & 0.75 & 196.53 & 1.72  \\
        \midrule
        \small{Ours-Full}  & 18.87 & 0.80 & 177.41 & 1.64  \\
        \bottomrule
    \end{tabular}
    \label{ablation}
        }
\end{table*}
\vspace{-5pt}

\section{Conclusion}
\label{sec:conclu}
    In this paper, we presented \ourmethod{}, an ``Observation-State-Reflection'' framework for camera-controllable video generation that explicitly enforces 4D spatiotemporal coherence. By grounding the model in Spatiotemporal Epipolar Causal Attention, our approach incorporates multi-view geometric constraints and temporal reasoning directly into the generation. Combined with reconstruction-driven geometric reflection and an active 4D retrieval strategy, the framework enables dynamic self-assessment and correction, effectively mitigating error accumulation.
Extensive experiments demonstrate that \ourmethod{} achieves state-of-the-art performance in synthesizing high-quality, free-view 4D scenes with strong geometric fidelity, temporal consistency, and trajectory accuracy.
% \clearpage

{\small
\bibliographystyle{unsrt}
\bibliography{main}

@String(CVPR= {IEEE Conf. Comput. Vis. Pattern Recog.})

@String(ICCV= {Int. Conf. Comput. Vis.})

@String(ECCV= {Eur. Conf. Comput. Vis.})

@String(CVPR  = {CVPR})

@String(ICCV  = {ICCV})

@String(ECCV  = {ECCV})

@article{he2024cameractrl,
  title={Cameractrl: Enabling camera control for text-to-video generation},
  author={He, Hao and Xu, Yinghao and Guo, Yuwei and Wetzstein, Gordon and Dai, Bo and Li, Hongsheng and Yang, Ceyuan},
  journal={arXiv preprint arXiv:2404.02101},
  year={2024}
}

@article{liu2025free4d,
  title={Free4D: Tuning-free 4D Scene Generation with Spatial-Temporal Consistency},
  author={Liu, Tianqi and Huang, Zihao and Chen, Zhaoxi and Wang, Guangcong and Hu, Shoukang and Shen, Liao and Sun, Huiqiang and Cao, Zhiguo and Li, Wei and Liu, Ziwei},
  journal={arXiv preprint arXiv:2503.20785},
  year={2025}
}

@inproceedings{ren2025gen3c,
  title={Gen3c: 3d-informed world-consistent video generation with precise camera control},
  author={Ren, Xuanchi and Shen, Tianchang and Huang, Jiahui and Ling, Huan and Lu, Yifan and Nimier-David, Merlin and Muller, Thomas and Keller, Alexander and Fidler, Sanja and Gao, Jun},
  booktitle={CVPR},
  pages={6121--6132},
  year={2025}
}

@inproceedings{wang2024motionctrl,
  title={Motionctrl: A unified and flexible motion controller for video generation},
  author={Wang, Zhouxia and Yuan, Ziyang and Wang, Xintao and Li, Yaowei and Chen, Tianshui and Xia, Menghan and Luo, Ping and Shan, Ying},
  booktitle={ACM SIGGRAPH},
  pages={1--11},
  year={2024}
}

@article{yao2025sv4d,
  title={Sv4d 2.0: Enhancing spatio-temporal consistency in multi-view video diffusion for high-quality 4d generation},
  author={Yao, Chun-Han and Xie, Yiming and Voleti, Vikram and Jiang, Huaizu and Jampani, Varun},
  journal={arXiv preprint arXiv:2503.16396},
  year={2025}
}

@article{yu2025trajectorycrafter,
  title={Trajectorycrafter: Redirecting camera trajectory for monocular videos via diffusion models},
  author={YU, Mark and Hu, Wenbo and Xing, Jinbo and Shan, Ying},
  journal={arXiv preprint arXiv:2503.05638},
  year={2025}
}

@article{yu2024viewcrafter,
  title={Viewcrafter: Taming video diffusion models for high-fidelity novel view synthesis},
  author={Yu, Wangbo and Xing, Jinbo and Yuan, Li and Hu, Wenbo and Li, Xiaoyu and Huang, Zhipeng and Gao, Xiangjun and Wong, Tien-Tsin and Shan, Ying and Tian, Yonghong},
  journal={arXiv preprint arXiv:2409.02048},
  year={2024}
}

@article{li2025vmem,
  title={VMem: Consistent Interactive Video Scene Generation with Surfel-Indexed View Memory},
  author={Li, Runjia and Torr, Philip and Vedaldi, Andrea and Jakab, Tomas},
  journal={arXiv preprint arXiv:2506.18903},
  year={2025}
}

@inproceedings{szymanowicz2025bolt3d,
  title={Bolt3d: Generating 3d scenes in seconds},
  author={Szymanowicz, Stanislaw and Zhang, Jason Y and Srinivasan, Pratul and Gao, Ruiqi and Brussee, Arthur and Holynski, Aleksander and Martin-Brualla, Ricardo and Barron, Jonathan T and Henzler, Philipp},
  booktitle={ICCV},
  pages={24846--24857},
  year={2025}
}

@article{bahmani2025lyra,
  title={Lyra: Generative 3d scene reconstruction via video diffusion model self-distillation},
  author={Bahmani, Sherwin and Shen, Tianchang and Ren, Jiawei and Huang, Jiahui and Jiang, Yifeng and Turki, Haithem and Tagliasacchi, Andrea and Lindell, David B and Gojcic, Zan and Fidler, Sanja and others},
  journal={arXiv preprint arXiv:2509.19296},
  year={2025}
}

@inproceedings{yu2025wonderworld,
  title={Wonderworld: Interactive 3d scene generation from a single image},
  author={Yu, Hong-Xing and Duan, Haoyi and Herrmann, Charles and Freeman, William T and Wu, Jiajun},
  booktitle={CVPR},
  pages={5916--5926},
  year={2025}
}

@article{song2025worldforge,
  title={Worldforge: Unlocking emergent 3d/4d generation in video diffusion model via training-free guidance},
  author={Song, Chenxi and Yang, Yanming and Zhao, Tong and Li, Ruibo and Zhang, Chi},
  journal={arXiv preprint arXiv:2509.15130},
  year={2025}
}

@article{wan2025wan,
  title={Wan: Open and advanced large-scale video generative models},
  author={Wan, Team and Wang, Ang and Ai, Baole and Wen, Bin and Mao, Chaojie and Xie, Chen-Wei and Chen, Di and Yu, Feiwu and Zhao, Haiming and Yang, Jianxiao and others},
  journal={arXiv preprint arXiv:2503.20314},
  year={2025}
}

@inproceedings{li2025realcam,
  title={Realcam-i2v: Real-world image-to-video generation with interactive complex camera control},
  author={Li, Teng and Zheng, Guangcong and Jiang, Rui and Zhan, Shuigen and Wu, Tao and Lu, Yehao and Lin, Yining and Deng, Chuanyun and Xiong, Yepan and Chen, Min and others},
  booktitle={ICCV},
  pages={28785--28796},
  year={2025}
}

@inproceedings{zhang2025spatialcrafter,
  title={SpatialCrafter: Unleashing the Imagination of Video Diffusion Models for Scene Reconstruction from Limited Observations},
  author={Zhang, Songchun and Xu, Huiyao and Guo, Sitong and Xie, Zhongwei and Bao, Hujun and Xu, Weiwei and Zou, Changqing},
  booktitle={ICCV},
  pages={27794--27805},
  year={2025}
}

@article{zhou2025stable,
  title={Stable virtual camera: Generative view synthesis with diffusion models},
  author={Zhou, Jensen and Gao, Hang and Voleti, Vikram and Vasishta, Aaryaman and Yao, Chun-Han and Boss, Mark and Torr, Philip and Rupprecht, Christian and Jampani, Varun},
  journal={arXiv preprint arXiv:2503.14489},
  year={2025}
}

@article{jiang2025anysplat,
  title={AnySplat: Feed-forward 3D Gaussian Splatting from Unconstrained Views},
  author={Jiang, Lihan and Mao, Yucheng and Xu, Linning and Lu, Tao and Ren, Kerui and Jin, Yichen and Xu, Xudong and Yu, Mulin and Pang, Jiangmiao and Zhao, Feng and others},
  journal={arXiv preprint arXiv:2505.23716},
  year={2025}
}

@article{lin2025movies,
  title={MoVieS: Motion-aware 4D dynamic view synthesis in one second},
  author={Lin, Chenguo and Lin, Yuchen and Pan, Panwang and Yu, Yifan and Yan, Honglei and Fragkiadaki, Katerina and Mu, Yadong},
  journal={arXiv preprint arXiv:2507.10065},
  year={2025}
}

@inproceedings{wang2025vggt,
  title={Vggt: Visual geometry grounded transformer},
  author={Wang, Jianyuan and Chen, Minghao and Karaev, Nikita and Vedaldi, Andrea and Rupprecht, Christian and Novotny, David},
  booktitle={CVPR},
  pages={5294--5306},
  year={2025}
}

@article{bai2025recammaster,
  title={Recammaster: Camera-controlled generative rendering from a single video},
  author={Bai, Jianhong and Xia, Menghan and Fu, Xiao and Wang, Xintao and Mu, Lianrui and Cao, Jinwen and Liu, Zuozhu and Hu, Haoji and Bai, Xiang and Wan, Pengfei and others},
  journal={arXiv preprint arXiv:2503.11647},
  year={2025}
}

@article{yang2024cogvideox,
  title={Cogvideox: Text-to-video diffusion models with an expert transformer},
  author={Yang, Zhuoyi and Teng, Jiayan and Zheng, Wendi and Ding, Ming and Huang, Shiyu and Xu, Jiazheng and Yang, Yuanming and Hong, Wenyi and Zhang, Xiaohan and Feng, Guanyu and others},
  journal={arXiv preprint arXiv:2408.06072},
  year={2024}
}

@article{ge2025arc,
  title={Arc-hunyuan-video-7b: Structured video comprehension of real-world shorts},
  author={Ge, Yuying and Ge, Yixiao and Li, Chen and Wang, Teng and Pu, Junfu and Li, Yizhuo and Qiu, Lu and Ma, Jin and Duan, Lisheng and Zuo, Xinyu and others},
  journal={arXiv preprint arXiv:2507.20939},
  year={2025}
}

@article{ma2025step,
  title={Step-video-t2v technical report: The practice, challenges, and future of video foundation model},
  author={Ma, Guoqing and Huang, Haoyang and Yan, Kun and Chen, Liangyu and Duan, Nan and Yin, Shengming and Wan, Changyi and Ming, Ranchen and Song, Xiaoniu and Chen, Xing and others},
  journal={arXiv preprint arXiv:2502.10248},
  year={2025}
}

@inproceedings{joseph2014nova,
  title={NOVA: QoE-driven optimization of DASH-based video delivery in networks},
  author={Joseph, Vinay and de Veciana, Gustavo},
  booktitle={INFOCOM},
  pages={82--90},
  year={2014},
  organization={IEEE}
}

@article{alhaija2025cosmos,
  title={Cosmos-transfer1: Conditional world generation with adaptive multimodal control},
  author={Alhaija, Hassan Abu and Alvarez, Jose and Bala, Maciej and Cai, Tiffany and Cao, Tianshi and Cha, Liz and Chen, Joshua and Chen, Mike and Ferroni, Francesco and Fidler, Sanja and others},
  journal={arXiv preprint arXiv:2503.14492},
  year={2025}
}

@article{li2024advances,
  title={Advances in 3d generation: A survey},
  author={Li, Xiaoyu and Zhang, Qi and Kang, Di and Cheng, Weihao and Gao, Yiming and Zhang, Jingbo and Liang, Zhihao and Liao, Jing and Cao, Yan-Pei and Shan, Ying},
  journal={arXiv preprint arXiv:2401.17807},
  year={2024}
}

@article{pages2018affordable,
  title={Affordable content creation for free-viewpoint video and VR/AR applications},
  author={Pages, Rafael and Amplianitis, Konstantinos and Monaghan, David and Ondrej, Jan and Smolic, Aljosa},
  journal={Journal of Visual Communication and Image Representation},
  volume={53},
  pages={192--201},
  year={2018},
  publisher={Elsevier}
}

@inproceedings{yang2024holodeck,
  title={Holodeck: Language guided generation of 3d embodied ai environments},
  author={Yang, Yue and Sun, Fan-Yun and Weihs, Luca and VanderBilt, Eli and Herrasti, Alvaro and Han, Winson and Wu, Jiajun and Haber, Nick and Krishna, Ranjay and Liu, Lingjie and others},
  booktitle={CVPR},
  pages={16227--16237},
  year={2024}
}

@article{wang2025embodiedgen,
  title={Embodiedgen: Towards a generative 3d world engine for embodied intelligence},
  author={Wang, Xinjie and Liu, Liu and Cao, Yu and Wu, Ruiqi and Qin, Wenkang and Wang, Dehui and Sui, Wei and Su, Zhizhong},
  journal={arXiv preprint arXiv:2506.10600},
  year={2025}
}

@article{zhou2024gala3d,
  title={Gala3d: Towards text-to-3d complex scene generation via layout-guided generative gaussian splatting},
  author={Zhou, Xiaoyu and Ran, Xingjian and Xiong, Yajiao and He, Jinlin and Lin, Zhiwei and Wang, Yongtao and Sun, Deqing and Yang, Ming-Hsuan},
  journal={arXiv preprint arXiv:2402.07207},
  year={2024}
}

@article{gao2024magicdrive3d,
  title={Magicdrive3d: Controllable 3d generation for any-view rendering in street scenes},
  author={Gao, Ruiyuan and Chen, Kai and Li, Zhihao and Hong, Lanqing and Li, Zhenguo and Xu, Qiang},
  journal={arXiv preprint arXiv:2405.14475},
  year={2024}
}

@inproceedings{lu2025infinicube,
  title={Infinicube: Unbounded and controllable dynamic 3d driving scene generation with world-guided video models},
  author={Lu, Yifan and Ren, Xuanchi and Yang, Jiawei and Shen, Tianchang and Wu, Zhangjie and Gao, Jun and Wang, Yue and Chen, Siheng and Chen, Mike and Fidler, Sanja and others},
  booktitle={ICCV},
  pages={27272--27283},
  year={2025}
}

@article{wu2024holodrive,
  title={Holodrive: Holistic 2d-3d multi-modal street scene generation for autonomous driving},
  author={Wu, Zehuan and Ni, Jingcheng and Wang, Xiaodong and Guo, Yuxin and Chen, Rui and Lu, Lewei and Dai, Jifeng and Xiong, Yuwen},
  journal={arXiv preprint arXiv:2412.01407},
  year={2024}
}

@inproceedings{geng2025motion,
  title={Motion prompting: Controlling video generation with motion trajectories},
  author={Geng, Daniel and Herrmann, Charles and Hur, Junhwa and Cole, Forrester and Zhang, Serena and Pfaff, Tobias and Lopez-Guevara, Tatiana and Aytar, Yusuf and Rubinstein, Michael and Sun, Chen and others},
  booktitle={CVPR},
  pages={1--12},
  year={2025}
}

@article{mildenhall2021nerf,
  title={Nerf: Representing scenes as neural radiance fields for view synthesis},
  author={Mildenhall, Ben and Srinivasan, Pratul P and Tancik, Matthew and Barron, Jonathan T and Ramamoorthi, Ravi and Ng, Ren},
  journal={Communications of the ACM},
  volume={65},
  number={1},
  pages={99--106},
  year={2021},
  publisher={ACM New York, NY, USA}
}

@inproceedings{yu2024mip,
  title={Mip-splatting: Alias-free 3d gaussian splatting},
  author={Yu, Zehao and Chen, Anpei and Huang, Binbin and Sattler, Torsten and Geiger, Andreas},
  booktitle={CVPR},
  pages={19447--19456},
  year={2024}
}

@article{keetha2025mapanything,
  title={MapAnything: Universal feed-forward metric 3D reconstruction},
  author={Keetha, Nikhil and Muller, Norman and Schonberger, Johannes and Porzi, Lorenzo and Zhang, Yuchen and Fischer, Tobias and Knapitsch, Arno and Zauss, Duncan and Weber, Ethan and Antunes, Nelson and others},
  journal={arXiv preprint arXiv:2509.13414},
  year={2025}
}

@inproceedings{po2024compositional,
  title={Compositional 3d scene generation using locally conditioned diffusion},
  author={Po, Ryan and Wetzstein, Gordon},
  booktitle={3DV},
  pages={651--663},
  year={2024}
}

@inproceedings{gao2024graphdreamer,
  title={Graphdreamer: Compositional 3d scene synthesis from scene graphs},
  author={Gao, Gege and Liu, Weiyang and Chen, Anpei and Geiger, Andreas and Scholkopf, Bernhard},
  booktitle={CVPR},
  pages={21295--21304},
  year={2024}
}

@article{kerbl20233d,
  title={3D Gaussian splatting for real-time radiance field rendering.},
  author={Kerbl, Bernhard and Kopanas, Georgios and Leimkuhler, Thomas and Drettakis, George},
  journal={ACM Trans. Graph.},
  volume={42},
  number={4},
  pages={139--1},
  year={2023}
}

@inproceedings{zhang2024gaussian,
  title={Gaussian in the wild: 3d gaussian splatting for unconstrained image collections},
  author={Zhang, Dongbin and Wang, Chuming and Wang, Weitao and Li, Peihao and Qin, Minghan and Wang, Haoqian},
  booktitle={ECCV},
  pages={341--359},
  year={2024},
  organization={Springer}
}

@article{wang20254d,
  title={4D-VGGT: A General Foundation Model with SpatioTemporal Awareness for Dynamic Scene Geometry Estimation},
  author={Wang, Haonan and Zhou, Hanyu and Liu, Haoyue and Yan, Luxin},
  journal={arXiv preprint arXiv:2511.18416},
  year={2025}
}

@article{chen2025deepverse,
  title={DeepVerse: 4D Autoregressive Video Generation as a World Model},
  author={Chen, Junyi and Zhu, Haoyi and He, Xianglong and Wang, Yifan and Zhou, Jianjun and Chang, Wenzheng and Zhou, Yang and Li, Zizun and Fu, Zhoujie and Pang, Jiangmiao and others},
  journal={arXiv preprint arXiv:2506.01103},
  year={2025}
}

@inproceedings{ling2024dl3dv,
  title={Dl3dv-10k: A large-scale scene dataset for deep learning-based 3d vision},
  author={Ling, Lu and Sheng, Yichen and Tu, Zhi and Zhao, Wentian and Xin, Cheng and Wan, Kun and Yu, Lantao and Guo, Qianyu and Yu, Zixun and Lu, Yawen and others},
  booktitle={CVPR},
  pages={22160--22169},
  year={2024}
}

@inproceedings{huang2024vbench,
  title={Vbench: Comprehensive benchmark suite for video generative models},
  author={Huang, Ziqi and He, Yinan and Yu, Jiashuo and Zhang, Fan and Si, Chenyang and Jiang, Yuming and Zhang, Yuanhan and Wu, Tianxing and Jin, Qingyang and Chanpaisit, Nattapol and others},
  booktitle={CVPR},
  pages={21807--21818},
  year={2024}
}

@article{zheng2025vbench,
  title={Vbench-2.0: Advancing video generation benchmark suite for intrinsic faithfulness},
  author={Zheng, Dian and Huang, Ziqi and Liu, Hongbo and Zou, Kai and He, Yinan and Zhang, Fan and Gu, Lulu and Zhang, Yuanhan and He, Jingwen and Zheng, Wei-Shi and others},
  journal={arXiv preprint arXiv:2503.21755},
  year={2025}
}

@misc{zhang2025joint,
      title={Joint 3D Geometry Reconstruction and Motion Generation for 4D Synthesis from a Single Image}, 
      author={Yanran Zhang and Ziyi Wang and Wenzhao Zheng and Zheng Zhu and Jie Zhou and Jiwen Lu},
      year={2025},
      eprint={2512.05044},
      archivePrefix={arXiv}
}

@article{lin2025depth,
  title={Depth anything 3: Recovering the visual space from any views},
  author={Lin, Haotong and Chen, Sili and Liew, Junhao and Chen, Donny Y and Li, Zhenyu and Shi, Guang and Feng, Jiashi and Kang, Bingyi},
  journal={arXiv preprint arXiv:2511.10647},
  year={2025}
}

@article{wu2025geometry,
  title={Geometry forcing: Marrying video diffusion and 3d representation for consistent world modeling},
  author={Wu, Haoyu and Wu, Diankun and He, Tianyu and Guo, Junliang and Ye, Yang and Duan, Yueqi and Bian, Jiang},
  journal={arXiv preprint arXiv:2507.07982},
  year={2025}
}

@article{dosovitskiy2016generating,
  title={Generating images with perceptual similarity metrics based on deep networks},
  author={Dosovitskiy, Alexey and Brox, Thomas},
  journal={NeurIPS},
  volume={29},
  year={2016}
}

@article{yang2026neoverse,
  title={NeoVerse: Enhancing 4D World Model with in-the-wild Monocular Videos},
  author={Yang, Yuxue and Fan, Lue and Shi, Ziqi and Peng, Junran and Wang, Feng and Zhang, Zhaoxiang},
  journal={arXiv preprint arXiv:2601.00393},
  year={2026}
}

@article{huang2026gen3r,
  title={Gen3R: 3D Scene Generation Meets Feed-Forward Reconstruction},
  author={Huang, Jiaxin and Yang, Yuanbo and Yang, Bangbang and Ma, Lin and Ma, Yuewen and Liao, Yiyi},
  journal={arXiv preprint arXiv:2601.04090},
  year={2026}
}
}

\end{document}